\documentclass{article}
\usepackage{algpseudocode}
\usepackage{spconf,amsmath,graphicx}
\usepackage{float}
\usepackage{makecell}
\usepackage{tabularx}
\usepackage{comment}
\usepackage{amssymb}
\usepackage{pifont}

\newcommand{\cmark}{\ding{51}}%
\newcommand{\xmark}{\ding{55}}%

\title{Revisiting Misalignment in Multispectral Pedestrian Detection: \\ A Language-Driven Approach for Cross-modal Alignment Fusion}
\name{Taeheon Kim$^{\dagger}$, Sangyun Chung$^{\dagger}$, Youngjoon Yu$^{\dagger}$, and Yong Man Ro$^*$\thanks{$^{\dagger}$Co-first authors: eetaekim@kaist.ac.kr, jelarum@kaist.ac.kr, greatday@kaist.ac.kr. $^*$Corresponding author: ymro@kaist.ac.kr. This work was conducted by Center for Applied Research in Artificial Intelligence (CARAI) grant funded by DAPA and ADD (UD230017TD).}}
\address{Integrated Vision and Language Lab, School of Electrical Engineering, KAIST, South Korea}

\begin{document}
%
\maketitle
\begin{abstract}
Multispectral pedestrian detection is a crucial component in various critical applications. However, a significant challenge arises due to the misalignment between these modalities, particularly under real-world conditions where data often appear heavily misaligned. Conventional methods developed on well-aligned or minimally misaligned datasets fail to address these discrepancies adequately. This paper introduces a new framework for multispectral pedestrian detection designed specifically to handle heavily misaligned datasets without the need for costly and complex traditional pre-processing calibration. By leveraging Large-scale Vision-Language Models (LVLM) for cross-modal semantic alignment, our approach seeks to enhance detection accuracy by aligning semantic information across the RGB and thermal domains. This method not only simplifies the operational requirements but also extends the practical usability of multispectral detection technologies in practical applications. 
\end{abstract}
\begin{keywords}
Multispectral Pedestrian Detection, Large-scale Vision-Language Models, Cross-modal Alignment
\end{keywords}
\section{INTRODUCTION}
\label{sec:intro}
\indent Multispectral pedestrian detection uses both RGB and infrared (thermal) images for pedestrian detection~\cite{hwang2015multispectral,kim2021uncertainty,yu2020investigating,kim2023multispectral, yu2021towards, park2023robust, park2021robust}. Compared to relying on just one modality detection~\cite{zong2023detrs,yu2022defending, kim2022defending,park2024integrating,park2024robust}, utilizing both modalities provide distinct advantages on pedestrian detection~\cite{kim2022map, kim2024causal, kim2024mscotdet}. However, positional misalignments between the modalities significantly degrade the models' performance. Generally, RGB and thermal cameras have different fields of view (FOV) and spatial distortions. In extreme cases, pedestrians may appear in completely different locations in the RGB and thermal images (Fig.1 (c)). Such scenarios challenge the model to fuse the information corresponding to the same pedestrian and puzzle the inference. \\
\indent 
Apart from these challenges, previous works on multispectral pedestrian detection have been developed on calibrated data, mostly containing well-aligned images or weakly aligned images, as illustrated in Fig.1 (a)-(b). This setting is often impractical because raw data are generally heavily misaligned. Also, obtaining calibrated data requires special pre-processing, and need to design special hardware such as beam splitters. Such constraints limit the practicability of current multispectral pedestrian detectors, making them difficult to be applied in the real world. \\ 
\indent We test conventional multispectral pedestrian detectors under uncalibrated data, as Fig.1 (c). We first collect uncalibrated data pairs by taking videos with two independent (non-calibrated) cameras: a mobile phone for RGB and a FLIR product for thermal. Illustrated in Fig.1 (c), the multispectral pedestrian detection model performs poorly on uncalibrated data, though it is evident from the human's perception. From these results, we can anticipate that current models fail to address heavily misaligned scenarios.\\
\begin{figure}[t!]
    \centering
    \includegraphics[width=0.8\linewidth]{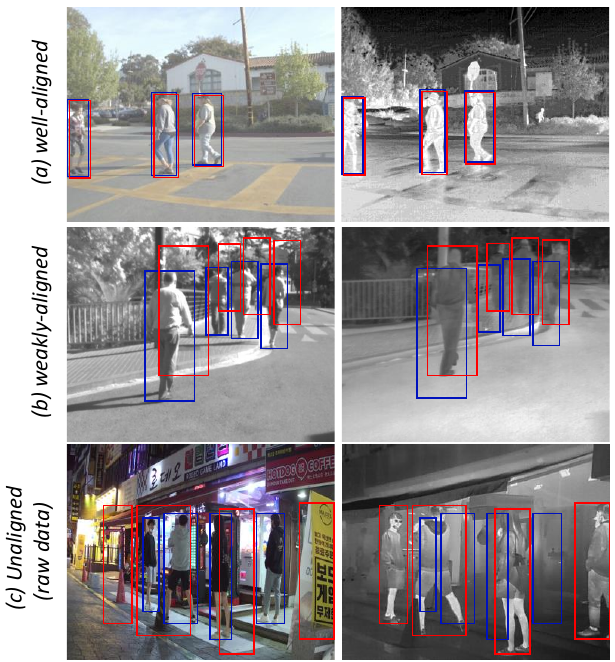}
    \noindent\caption{Visualized examples of well-aligned, weakly-aligned, and unaligned (raw data) scenarios in multispectral pedestrian detection. Unaligned pose a challenge as the detected persons in RGB (blue box) and in thermal (red box) does not overlap.
    }
    \vspace{-0.5cm}
    \label{fig:1}
\end{figure}
\indent \hspace{-0.3cm} Our goal is to perform accurate multispectral pedestrian detection on heavily misaligned scenarios. If this is possible, we no longer have to use expensive calibrating devices and pre-processing data. And we can directly perform multispectral pedestrian detection with raw data taken from uncalibrated cameras. To ensure accurate detection, the model must first identify the same individual across both modalities and then integrate the corresponding information for detection. \\
\indent Toward this goal, we first investigate human perception abilities. Humans are adept at recognizing individuals across different modalities by understanding the scenes. This includes global contexts, such as pedestrians' position and relationship with the environment. Moreover, humans can understand local contexts such as each pedestrian's shape, movement, and appearance. Using these contextual cues, humans can comprehend to fuse the correct pedestrian information even from misaligned scenarios. \\
\indent Motivated by the humans' ability, we propose a novel cross-modal semantic alignment method to handle the aforementioned challenging misalignment problems in multispectral pedestrian detection. To be specific, our method consists of three parts: constructing positional graphs, embedding appearance information, and prediction with large-scale vision-language models. First the single-modal detection is conducted in each modality using the Co-DETR model~\cite{zong2023detrs} trained on the FLIR\_ADAS dataset. This process yields bounding box coordinates of pedestrians. These coordinates are used to construct a graph where nodes denote instance coordinates and edges denote distances between instances.\\ \indent Second, in the embedding appearance information, we recognize individuals by their unique physical contours and characteristic movements, which can be identifiable even when visual appearances differ between modalities. To leverage this, appearance information for each individual is obtained using a Large-scale Vision-Language Models (LVLMs). To address common hallucination issues in LVLMs and ensure accurate descriptions, established visual pre-processing methods are employed. 

The main contributions are summarized as follows:
\begin{itemize}
\item We integrate the Large-scale Vision-Language Models (LVLM) to tackle challenging misalignment problems that are not readily resolved with conventional image processing methods.
\item We thoroughly validated the proposed approach using the challenging misalignment datasets.
\end{itemize}
\section{Related Work}
\noindent \textbf{Misalignments in Multispectral Pedestrian Detection.} \vspace{0.2cm} \\ 
To solve the misalignment problems, camera calibration techniques and image registration algorithms have been developed. They pre-process the raw data to be spatially aligned. Camera calibration techniques physically align the two image domains using special devices such as beam splitters~\cite{hwang2015multispectral}  or checkerboards. Yet, these techniques are expensive, labor-intensive, and designed on specific camera hardware. Image registration algorithms~\cite{brown1992survey,dawn2010remote,maintz1998survey,torabi2012iterative} geometrically align two images. But their performance significantly drops when the misalignment is above a certain degree.\\
\indent Other methods try to perform feature-level alignment by predicting spatial offsets~\cite{chen2023attentive, zhang2019weakly}. Other methods often fuse RGB and thermal using illumination-based weighting strategies~\cite{li2019illumination,zhou2020improving}, uncertainty estimates~\cite{kim2021uncertainty}, or confidence
scores~\cite{chen2022multimodal, kim2024mscotdet}. However, these techniques operate under the assumption that the same pedestrian will appear in ROIs of largely overlapping areas within the different modalities. Therefore, these methods cannot be applied to heavily misaligned scenarios.
\begin{figure*}[t!]
    \centering
    \includegraphics[width=\textwidth]{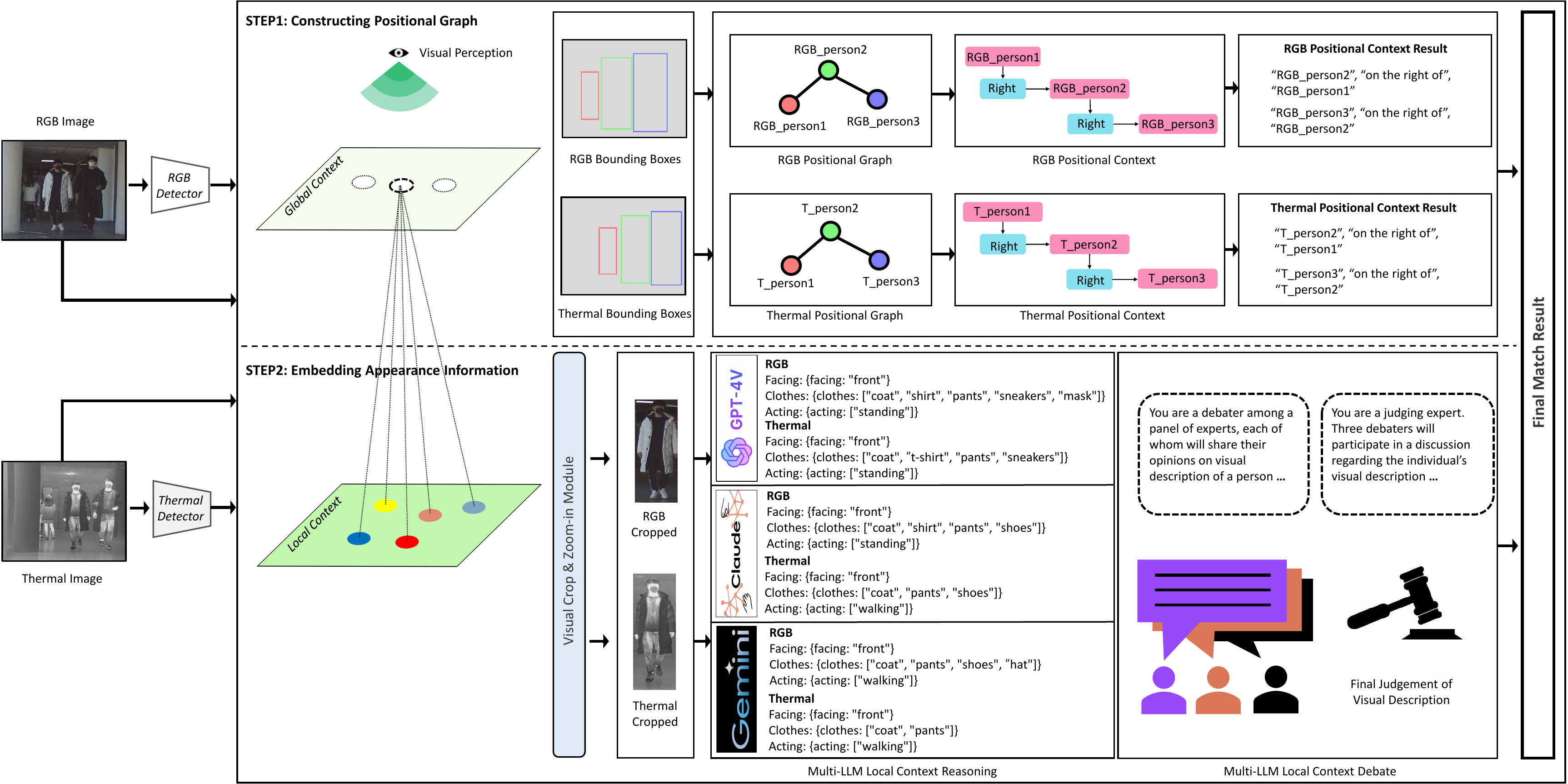}
    \noindent\caption{Overall architecture of the proposed cross-modal semantic alignment fusion method.
    }
    \vspace{-0.2cm}
    \label{fig:2}
\end{figure*}
\section{PROPOSED METHOD}
\label{sec:method}

\subsection{Constructing Positional Graphs}
\indent Humans can analyze the scenes and extract semantic contexts useful for pedestrian detection using global and local contexts of the scene. Specifically, humans can recognize pedestrians at a glance by their positional location relative to surrounding objects or landmarks. Such positional relations can be treated as global contexts. For example, seeing someone standing in a familiar room or on a specific street corner can trigger recognition, even if the visual characteristics differ between modalities. Observing how an individual interacts with objects, other people, or their surroundings can provide valuable context for recognition. Since both types of images are based on the physical arrangement of space, the relative positions and distances of people will appear similarly. This property can be utilized to implement robust person detection under misalignment conditions.  \\ 
\indent We represent the positional relationships between pedestrian coordinates as a graph, consisting of nodes and edges. Nodes represent the coordinates of each pedestrian, while edges represent the distances between pedestrians. The edges of the graph are connected to the closest person, and each node is limited to having a maximum of two edges. We construct a Minimum Spanning Tree (MST) to connect the nodes on the optimal connection structure. Using Kruskal's algorithm we connect all the nodes with the minimum possible Euclidian distance and ensure that no cycles are formed. This process is described in Figure 2. As a result, two positional graphs are extracted as below.
\subsection{Embedding Appearance Information}
\indent Next, we add the local context to the positional graphs, which involve appearance descriptions of each person. This could involve details like clothing, facing direction, or other visual characteristics. Each person has a unique body shape and silhouette, which remains relatively consistent across different imaging modalities. And the movement of a person is highly characteristic and can be a powerful cue for recognition. \\
\indent Building on this approach, we acquire appearance information for each individual using a Large-scale Vision-Language Model (LVLM). To mitigate the common occurrence of hallucinations in LVLMs and obtain accurate descriptions, we utilize two strategies. First, we adapt visual pre-processing methods from existing approaches. This method involves initially zooming in on the region of interest surrounding the region of interest (ROI) and then cropping the zoomed-in image before feeding it into the LVLM as input. The prompt being used is as follows. \\
\textit{Input prompt: ``Given an RGB image and a thermal image, generate a textual description of the appearance of each individual in the scene. The description should include information about clothing, accessories, hairstyle, and any other visible attributes..."}\\
\indent Second, by comparing responses from multiple LVLM models (such as GPT-4~\cite{achiam2023gpt}, Gemini~\cite{reid2024gemini}, and Claude 2), we cross-validate information to further mitigate hallucination. If all models agree on an answer, it increases confidence in its accuracy. Conversely, reasoning from LVLMs can sometimes produce incorrect answers, and recent studies have also demonstrated that LVLMs struggle to self-correct their responses without external feedback. If they provide differing answers, it flags a need for further verification. To address this issue, we have incorporated a debate scheme to synthesize responses from different LVLMs. An example of what a debater’s prompt might look like:\\
\textit{Input prompt: ``You are a debater among a panel of expert, each of whom will share their opinions on visual description of a person. Please share your opinions in brief..."}\\
\indent Moreover, we characterize the role of the judge to deduce the conclusive answer from the aggregate debate history. A sample prompt for the judge is delineated as:\\
\textit{Input prompt: ``You are a judging expert. Three debaters will participate in a discussion regarding the individual's visual description. Once the debate is over, it will be your responsibility to decide which visual description seems most reasonable one based on the debate content."}
\\
\indent As multiple LVLMs are involved but we only need one judge, we choose the generally most powerful LVLM, GPT-4, to play the role of judge.\\
\indent We add these textual descriptions to each node of the graph. The final graph is used to output matching results.
\subsection{Prediction with LLMs}
Finally, we prompt the LLM to make predictions. The designed prompt first generates rationales and outputs the matching results.\\
\textit{Input prompt: ``This labeling system is designed to assist you in matching the people in misaligned thermal images and RGB images... Please answer in the following format. }\\
\textit{Rationale: write your rational\\
Matching result: (RGB\_person1 : T\_person1, RGB\_person2 : T\_person2, ...)" }\\
The final output is the matching results. In the following section, we explain the evaluation process of our cross-modal semantic alignment model with our newly proposed metrics.

\begin{figure*}[t!]
    \centering
    \includegraphics[width=1.0\textwidth]{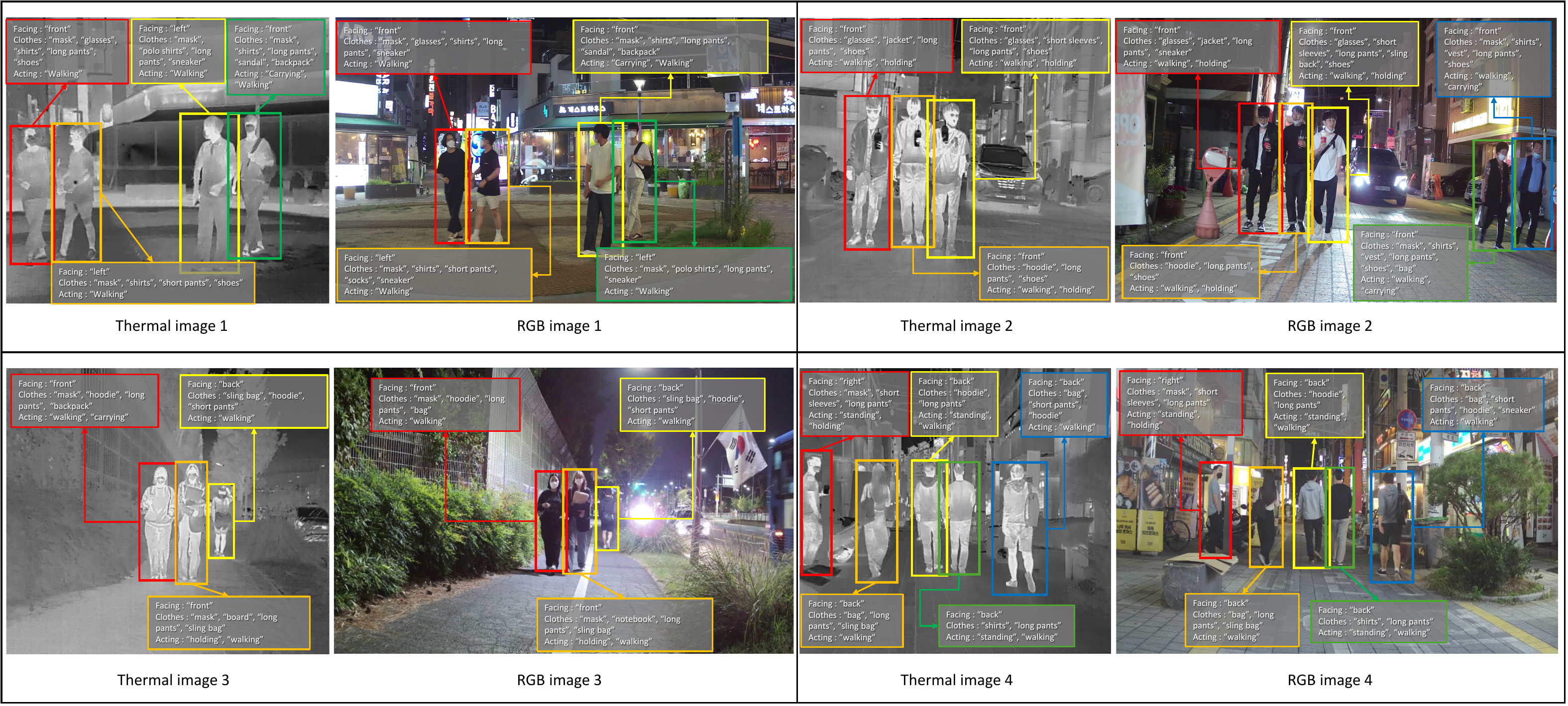}
    \noindent\caption{Visualized qualitative results: the same color means the same identity of a matched person across two different sensors.
    }
    \label{fig:3}
\end{figure*}

\section{EXPERIMENTS}
\subsection{Implementation Details}
We conduct experiments on two different heavily misaligned RGB-thermal multispectral dataset. One is modified from public FLIR\_ADAS dataset, and the other is our collected dataset. Each dataset contains a hundred challenging misalignment RGB-thermal unaligned pairs respectively. For the comparison, we use the baseline model as ProbEn~\cite{chen2022multimodal}, which is recently proposed late-fusion multispectral pedestrian detection model. We compare this baseline model with our proposed cross-modal alignment fusion method due to their well-known robustness against weakly-aligned dataset~\cite{chen2022multimodal}. 

\subsection{Quantitative Results}
To validate our proposed method in the challenging misalignment scenarios, we propose a new metric called alignment error rate (AER) to measure the degree of misalignment severity in multispectral pedestrian detection. This is because there is no prior work on evaluating heavily misaligned RGB-thermal pair. The AER is defined as follows:
\begin{equation}
\textnormal{AER} = \frac{\text{\# of mismatched RGB-T person pairs}}{\text{\# of total RGB-T person pairs}}.
\label{eq11}
\end{equation}

\noindent Table 1 shows that our proposed cross-modal alignment fusion method outperforms the baseline not only in terms of the AP score but also the newly adopted AER alignment score.


\subsection{Qualitative Results}
Fig. 3 depicts four detection results, validating that our framework can recognize discrepancies and perform multispectral pedestrian detection by matching the same person. Specifically, the fusion is achieved by identifying the same individual through their positional relationships and visual attributes. As a result, our method proves effective in unaligned scenarios.

\begin{table}[t!]
\centering
\renewcommand{\arraystretch}{1.1}
\caption{\textbf{Quantitative Results.}}
\vspace{0.2cm}
\resizebox{1.0\linewidth}{!}{
\begin{tabular}{clccccccc}
\hline
Dataset  &  & \multicolumn{3}{c}{FLIR Challenging DB} &  & \multicolumn{3}{c}{Collected Challenging DB} \\ \cline{1-1} \cline{3-5} \cline{7-9}
Model    &  & LLM Debate     & AP($\uparrow$)       & AER($\downarrow$)      &  & LLM Debate       & AP($\uparrow$)         & AER($\downarrow$)       \\ \cline{3-5} \cline{7-9} 
Baseline &  & \xmark           & 61.6     & 82.3     &  & \xmark                & 75.6       & 85.8      \\
Ours &  & \xmark               & 72.0     & 21.3     &  & \xmark                & 80.4       & 29.8      \\
Ours &  & \cmark              & \textbf{74.9}     & \textbf{8.8}      &  & \cmark                & \textbf{85.5}       & \textbf{22.0}      \\ \hline
\end{tabular}}
\end{table}
\vspace{-0.2cm}

\section{CONCLUSION}
\label{sec:conc}

This paper introduces an innovative framework for multispectral pedestrian detection, specifically addressing the challenges of heavy misalignment between RGB and thermal images in practical applications. Unlike conventional methods that rely on costly and complex pre-processing to align images, our approach utilizes semantic alignment through large language models to improve alignment and detection accuracy directly on raw, uncalibrated data. This advancement significantly reduces the dependency on specialized hardware and opens new possibilities for deploying multispectral pedestrian detection in real-world systems. Future work will focus on refining the semantic alignment techniques and expanding the framework's applicability to other multispectral imaging tasks beyond pedestrian detection.

\bibliographystyle{IEEEbib}
\bibliography{strings,refs}

\begin{thebibliography}{10}

\bibitem{hwang2015multispectral}
Soonmin Hwang, Jaesik Park, Namil Kim, Yukyung Choi, and In~So~Kweon,
\newblock ``Multispectral pedestrian detection: Benchmark dataset and baseline,''
\newblock in {\em Proceedings of the IEEE conference on computer vision and pattern recognition}, 2015, pp. 1037--1045.

\bibitem{kim2021uncertainty}
Jung~Uk Kim, Sungjune Park, and Yong~Man Ro,
\newblock ``Uncertainty-guided cross-modal learning for robust multispectral pedestrian detection,''
\newblock {\em IEEE Transactions on Circuits and Systems for Video Technology}, vol. 32, no. 3, pp. 1510--1523, 2021.

\bibitem{yu2020investigating}
Youngjoon Yu, Hong~Joo Lee, Byeong~Cheon Kim, Jung~Uk Kim, and Yong~Man Ro,
\newblock ``Investigating vulnerability to adversarial examples on multimodal data fusion in deep learning,''
\newblock {\em arXiv preprint arXiv:2005.10987}, 2020.

\bibitem{kim2023multispectral}
Taeheon Kim, Youngjoon Yu, and Yong~Man Ro,
\newblock ``Multispectral invisible coating: laminated visible-thermal physical attack against multispectral object detectors using transparent low-e films,''
\newblock in {\em Proceedings of the AAAI Conference on Artificial Intelligence}, 2023, vol.~37, pp. 1151--1159.

\bibitem{yu2021towards}
Youngjoon Yu, Hong~Joo Lee, Byeong~Cheon Kim, Jung~Uk Kim, and Yong~Man Ro,
\newblock ``Towards robust training of multi-sensor data fusion network against adversarial examples in semantic segmentation,''
\newblock in {\em ICASSP 2021-2021 IEEE International Conference on Acoustics, Speech and Signal Processing (ICASSP)}. IEEE, 2021, pp. 4710--4714.

\bibitem{park2023robust}
Sungjune Park, Jung~Uk Kim, Jin~Mo Song, and Yong~Man Ro,
\newblock ``Robust multispectral pedestrian detection via spectral position-free feature mapping,''
\newblock in {\em 2023 IEEE International Conference on Image Processing (ICIP)}. IEEE, 2023, pp. 1795--1799.

\bibitem{park2021robust}
Sungjune Park, Jung~Uk Kim, Yeon~Gyun Kim, Sang-Keun Moon, and Yong~Man Ro,
\newblock ``Robust multispectral pedestrian detection via uncertainty-aware cross-modal learning,''
\newblock in {\em MultiMedia Modeling: 27th International Conference, MMM 2021, Prague, Czech Republic, June 22--24, 2021, Proceedings, Part I 27}. Springer, 2021, pp. 391--402.

\bibitem{zong2023detrs}
Zhuofan Zong, Guanglu Song, and Yu~Liu,
\newblock ``Detrs with collaborative hybrid assignments training,''
\newblock in {\em Proceedings of the IEEE/CVF international conference on computer vision}, 2023, pp. 6748--6758.

\bibitem{yu2022defending}
Youngjoon Yu, Hong~Joo Lee, Hakmin Lee, and Yong~Man Ro,
\newblock ``Defending person detection against adversarial patch attack by using universal defensive frame,''
\newblock {\em IEEE Transactions on Image Processing}, vol. 31, pp. 6976--6990, 2022.

\bibitem{kim2022defending}
Taeheon Kim, Youngjoon Yu, and Yong~Man Ro,
\newblock ``Defending physical adversarial attack on object detection via adversarial patch-feature energy,''
\newblock in {\em Proceedings of the 30th ACM International Conference on Multimedia}, 2022, pp. 1905--1913.

\bibitem{park2024integrating}
Sungjune Park, Hyunjun Kim, and Yong~Man Ro,
\newblock ``Integrating language-derived appearance elements with visual cues in pedestrian detection,''
\newblock {\em IEEE Transactions on Circuits and Systems for Video Technology}, 2024.

\bibitem{park2024robust}
Sungjune Park, Hyunjun Kim, and Yong~Man Ro,
\newblock ``Robust pedestrian detection via constructing versatile pedestrian knowledge bank,''
\newblock {\em Pattern Recognition}, p. 110539, 2024.

\bibitem{kim2022map}
Taeheon Kim, Hong~Joo Lee, and Yong~Man Ro,
\newblock ``Map: Multispectral adversarial patch to attack person detection,''
\newblock in {\em ICASSP 2022-2022 IEEE International Conference on Acoustics, Speech and Signal Processing (ICASSP)}. IEEE, 2022, pp. 4853--4857.

\bibitem{kim2024causal}
Taeheon Kim, Sebin Shin, Youngjoon Yu, Hak~Gu Kim, and Yong~Man Ro,
\newblock ``Causal mode multiplexer: A novel framework for unbiased multispectral pedestrian detection,''
\newblock in {\em Proceedings of the IEEE/CVF Conference on Computer Vision and Pattern Recognition}, 2024, pp. 26784--26793.

\bibitem{kim2024mscotdet}
Taeheon Kim, Sangyun Chung, Damin Yeom, Youngjoon Yu, Hak~Gu Kim, and Yong~Man Ro,
\newblock ``Mscotdet: Language-driven multi-modal fusion for improved multispectral pedestrian detection,''
\newblock {\em arXiv preprint arXiv:2403.15209}, 2024.

\bibitem{brown1992survey}
Lisa~Gottesfeld Brown,
\newblock ``A survey of image registration techniques,''
\newblock {\em ACM computing surveys (CSUR)}, vol. 24, no. 4, pp. 325--376, 1992.

\bibitem{dawn2010remote}
Suma Dawn, Vikas Saxena, and Bhudev Sharma,
\newblock ``Remote sensing image registration techniques: A survey,''
\newblock in {\em Image and Signal Processing: 4th International Conference, ICISP 2010, Trois-Rivi{\`e}res, QC, Canada, June 30-July 2, 2010. Proceedings 4}. Springer, 2010, pp. 103--112.

\bibitem{maintz1998survey}
JB~Antoine Maintz and Max~A Viergever,
\newblock ``A survey of medical image registration,''
\newblock {\em Medical image analysis}, vol. 2, no. 1, pp. 1--36, 1998.

\bibitem{torabi2012iterative}
Atousa Torabi, Guillaume Mass{\'e}, and Guillaume-Alexandre Bilodeau,
\newblock ``An iterative integrated framework for thermal--visible image registration, sensor fusion, and people tracking for video surveillance applications,''
\newblock {\em Computer Vision and Image Understanding}, vol. 116, no. 2, pp. 210--221, 2012.

\bibitem{chen2023attentive}
Nuo Chen, Jin Xie, Jing Nie, Jiale Cao, Zhuang Shao, and Yanwei Pang,
\newblock ``Attentive alignment network for multispectral pedestrian detection,''
\newblock in {\em Proceedings of the 31st ACM international conference on multimedia}, 2023, pp. 3787--3795.

\bibitem{zhang2019weakly}
Lu~Zhang, Xiangyu Zhu, Xiangyu Chen, Xu~Yang, Zhen Lei, and Zhiyong Liu,
\newblock ``Weakly aligned cross-modal learning for multispectral pedestrian detection,''
\newblock in {\em Proceedings of the IEEE/CVF international conference on computer vision}, 2019, pp. 5127--5137.

\bibitem{li2019illumination}
Chengyang Li, Dan Song, Ruofeng Tong, and Min Tang,
\newblock ``Illumination-aware faster r-cnn for robust multispectral pedestrian detection,''
\newblock {\em Pattern Recognition}, vol. 85, pp. 161--171, 2019.

\bibitem{zhou2020improving}
Kailai Zhou, Linsen Chen, and Xun Cao,
\newblock ``Improving multispectral pedestrian detection by addressing modality imbalance problems,''
\newblock in {\em Computer Vision--ECCV 2020: 16th European Conference, Glasgow, UK, August 23--28, 2020, Proceedings, Part XVIII 16}. Springer, 2020, pp. 787--803.

\bibitem{chen2022multimodal}
Yi-Ting Chen, Jinghao Shi, Zelin Ye, Christoph Mertz, Deva Ramanan, and Shu Kong,
\newblock ``Multimodal object detection via probabilistic ensembling,''
\newblock in {\em European Conference on Computer Vision}. Springer, 2022, pp. 139--158.

\bibitem{achiam2023gpt}
Josh Achiam, Steven Adler, Sandhini Agarwal, Lama Ahmad, Ilge Akkaya, Florencia~Leoni Aleman, Diogo Almeida, Janko Altenschmidt, Sam Altman, Shyamal Anadkat, et~al.,
\newblock ``Gpt-4 technical report,''
\newblock {\em arXiv preprint arXiv:2303.08774}, 2023.

\bibitem{reid2024gemini}
Machel Reid, Nikolay Savinov, Denis Teplyashin, Dmitry Lepikhin, Timothy Lillicrap, Jean-baptiste Alayrac, Radu Soricut, Angeliki Lazaridou, Orhan Firat, Julian Schrittwieser, et~al.,
\newblock ``Gemini 1.5: Unlocking multimodal understanding across millions of tokens of context,''
\newblock {\em arXiv preprint arXiv:2403.05530}, 2024.

\end{thebibliography}

\end{document}